\documentclass{article}

\usepackage[preprint]{neurips_2026}

\usepackage[utf8]{inputenc}
\usepackage[T1]{fontenc}
\usepackage{hyperref}
\usepackage{url}
\usepackage{booktabs}
\usepackage{amsfonts}
\usepackage{amsmath}
\usepackage{amssymb}
\usepackage{mathtools}
\usepackage{nicefrac}
\usepackage{microtype}
\usepackage{xcolor}
\usepackage{graphicx}
\usepackage{subcaption}
\usepackage{multirow}
\usepackage{array}
\usepackage{placeins}
\usepackage{enumitem}
\usepackage{placeins}
\usepackage{float}
\usepackage{hyperref}       % hyperlinks
\usepackage{url}            % simple URL typesetting
\usepackage{booktabs}       % professional-quality tables
\usepackage{amsfonts}       % blackboard math symbols
\usepackage{nicefrac}       % compact symbols for 1/2, etc.
\usepackage{microtype}      % microtypography
\usepackage{graphicx}
\usepackage{booktabs}
\usepackage{tabularx}
\usepackage{longtable}
\usepackage{amsmath}
\usepackage{multirow}
\usepackage{pdflscape}

\providecommand{\scoreerr}[1]{{\tiny $\pm$#1}}

\begin{document}
\title{PPG-Bench: A Unified Benchmark for PPG Foundation Models}

\author{%
	Arvasu Kulkarni\\
	Mannas AI\\
	\texttt{arvasu@mannas.ai}
	\And
	Mahir Jain\\
	Mannas AI\\
	\texttt{mahir@mannas.ai}
	\And
	Aditya Ray Mishra\\
	Mannas AI\\
	\texttt{aditya@mannas.ai}
	\And
	Raj Singh Bani\\
	Mannas AI\\
	\texttt{raj@mannas.ai}
	\And
	Sandeep Singh\\
	Mannas AI\\
	\texttt{sandeep@mannas.ai}
	\And
	Siddharth Panwar\\
	Mannas AI\\
	\texttt{siddharth@mannas.ai}
}

\maketitle

	\begin{abstract}
	Photoplethysmography (PPG) is increasingly recorded by wearable and clinical devices at enormous scale, and a growing number of PPG foundation models (FMs) have been developed to compute reusable representations for various downstream tasks. These models are difficult to compare: different evaluations use different datasets, task definitions, preprocessing, downstream heads, hyperparameter tuning, and baselines. We present PPG-Bench, a unifying open-source benchmark evaluating five public PPG FMs across nine open-access datasets and 26 tasks spanning cardiovascular state estimation, activity, sleep, screening, and identification. We evaluate models under four downstream regimes of increasing adaptability (linear probing, random forests, frozen MLP probes, and end-to-end finetuning), with model-specific hyperparameter optimization and subject-disjoint cross-validation. We also evaluate against three controls: random/constant baseline, a deterministic pulse-rate-variability (PRV) feature set, and a generic time-series FM (MOMENT). Across all 68 task–regime combinations, 35.3\% show no meaningful FM advantage over the random/constant baseline, and in 83.8\% no PPG-specific FM outperforms both PRV and MOMENT. We also find that added downstream head capacity generally improves performance, but can cause significant degradation if the model is performing at chance/constant level. Current PPG FMs therefore cannot yet claim to provide generalised and transferrable representations of PPG signals, and future evaluation in this field should include checks against simple, non-learned controls. Code for the benchmark and instructions to evaluate more models can be found here - https://github.com/mannas-ai/ppg-bench.
	\end{abstract}
	
	\section{Introduction}
	
	Photoplethysmography (PPG) is widely recorded by wearable and clinical devices and supports applications spanning cardiovascular monitoring, activity monitoring, sleep staging, and clinical screening~\citep{charlton2023roadmap,loh2022ppggeneral}. This has motivated research into PPG foundation models (FMs) intended to provide reusable representations across downstream tasks~\citep{abbaspourazad2024apple,guo2026sigmappg,nie2026anyppg,pillai2025papagei,saha2025pulseppg}.
	
	However, current PPG FMs are difficult to compare. Their evaluations differ in
	datasets, task definitions, preprocessing, downstream heads, finetuning
	procedures, hyperparameter optimization, and baselines. Comparisons
	between PPG FMs also do not establish whether learned representations
	improve over conventional physiological features or whether PPG-specific
	pretraining adds value beyond generic time-series pretraining.
	
	To address these issues, we introduce \textbf{PPG-Bench}, an open-source benchmark that tests five public PPG
	FMs across nine open-access datasets and 17 primary + 9 secondary tasks. Models are evaluated
	using linear probes, random forests, frozen MLP probes, and end-to-end
	finetuning, with model-specific hyperparameter optimization and fully
	subject-disjoint cross-validation. We compare them against a random/constant
	baseline, a deterministic pulse-rate-variability (PRV) representation, and
	MOMENT, a generic time-series FM~\citep{goswami2024moment}.
	
	Across 68 task--regime combinations (regime meaning the type of evaluation method used), 24 (35.3\%) show no meaningful PPG-FM improvement over the random chance baseline, and in 57 (83.8\%) no PPG-specific FM outperforms both PRV and MOMENT. These results show that current PPG FMs provide inconsistent gains over simpler alternatives and motivate common, controlled evaluation.
	
	\section{Related Work}
	
	A growing number of foundation models for photoplethysmography have been proposed in recent years, including the public models evaluated in this work~\citep{nie2026anyppg,pillai2025papagei,saha2025pulseppg,guo2026sigmappg}. These models are commonly evaluated across multiple downstream tasks and datasets. However, the evaluation protocol is typically defined within each model paper, so dataset selection, task formulation, preprocessing, downstream heads, finetuning procedures, hyperparameter optimization, and comparator models can differ substantially across studies. PPG-Bench instead evaluates released checkpoints under a common protocol.
	
	Recent studies have introduced broader benchmark and benchmark-adjacent evaluations for biosignal foundation models. In the PPG domain, one study compares a specialist PPG foundation model with a general time-series foundation model across a large collection of datasets and downstream tasks~\citep{kataria2025genspec}. Another benchmark evaluates multiple physiological foundation models on synchronized ECG and PPG across numerous clinical prediction tasks derived from a large clinical dataset~\citep{gustafsson2026mcmed}. Dataset-release papers can also include multi-model benchmark evaluations; for example, BioMTL compares several models on a new PPG dataset~\citep{zhang2026biomtl}. These studies provide useful standardized comparisons, but differ from PPG-Bench in either the number of models considered or the use of a single underlying dataset rather than multiple independently collected PPG datasets. To our knowledge, PPG-Bench is the first independent PPG FM benchmark that covers both multiple foundation models and multiple datasets.
	
	Similar benchmarking efforts have emerged in other biosignal modalities, which have seen an uptick in FM development. EEG-FM-Bench standardizes evaluation of EEG foundation models across multiple datasets and downstream paradigms~\citep{xiong2025eegfmbench}, while recent ECG benchmark work compares ECG-specific and general-purpose foundation models across multiple prediction tasks~\citep{almasud2026ecg}.
	
	Benchmark rankings can be sensitive to hyperparameter and training recipe choices. Additionally, prior work has shown that model ranking based on cross-validation can introduce significant bias due to repeated hyperparameter trials over the same CV folds. Nested cross-validation provides a way to separate model selection from evaluation, although practical alternatives can offer similar model-selection quality at lower computational cost in some settings~\citep{cawley2010, wainer2021nestedcv}. These considerations motivate the model-specific HPO and subject-disjoint nested CV evaluation procedures used in PPG-Bench.
	
	Taken together, existing efforts either compare a limited set of PPG models across many datasets or evaluate multiple models on a single dataset. PPG-Bench complements these studies by providing a common, extensible evaluation framework for multiple public PPG-specific FMs across independently collected datasets, with standardized downstream adaptation and hyperparameter optimization. The open-source nature of this work will allow teams developing PPG-FMs to report standardized evaluations of their models, and would also allow researchers to add datasets.
	
	\section{Datasets, Tasks, and Preprocessing}
	We use nine open-access datasets covering 17 primary and nine secondary tasks (Appendix~\ref{sec:secondary} and Table~\ref{tab:dataset-overview}). They span cardiovascular estimation, activity, affect, sleep, screening, and identification, while varying in recording environment, hardware, and sensor location. This is done to reflect the fact that FMs are expected to perform well across a wide range of tasks and settings.
	
	\paragraph{Preprocessing}
	PPG-Bench applies minimal preprocessing and otherwise preserves each model's native preprocessing pipeline as faithfully as possible, to ensure released checkpoints do not encounter OOD (out-of-distribution) data. Source data are converted into a common segmented representation and resampled as required, while model-specific operations such as normalization, filtering, cropping, padding, or feature construction follow the preprocessing used by the corresponding pretrained model.
	
\paragraph{Tasks}
We carefully consider \textit{which tasks} should be included in a PPG-FM benchmark and how those tasks should be defined. We identify two cases where the inclusion must be critically evaluated.

\begin{enumerate}
	\item \textbf{Externally available or low-incremental-value targets.}
	Some targets may be predictable from PPG but are already known without physiological inference. For example, one FM evaluates pregnancy-stage classification on the nuMoM2B dataset using recordings from early and later clinic visits~\citep{haas2015numo2b,pillai2025papagei}. Coarse pregnancy stage is already known to most subjects, so predicting it from PPG has limited standalone value. Age and BMI estimation tasks raise a similar issue, as both are normally directly available. Such targets can still be scientifically useful. Age prediction, for example, could potentially capture differences between chronological and physiological age, but this requires sufficiently accurate prediction (which has not yet been demonstrated by FMs) and evidence that the estimate reflects physiological ageing. However, we retain some of these tasks for comparability with prior work, while not treating them as strong evidence of FM capability.
	
	\item \textbf{Tasks constructed from ambiguous labels.}
	Some datasets provide continuous scores rather than predefined classes, requiring FM researchers to decide how the downstream task should be constructed. These choices can substantially affect both task difficulty and interpretation. ECSMP~\citep{gao2021ecsmp}, for example, provides SDS~\citep{zung1962sds} and POMS-TMD~\citep{mcnair1971poms} questionnaire scores. One FM converts TMD into a binary task using the dataset median as a threshold~\citep{pillai2025papagei}. Because this median is not an established clinical threshold, the resulting classes are specific to that cohort. PPG-Bench instead uses the SDS score with its established threshold of 50 for clinically relevant depression classification. Similarly, previous work converts WESAD valence and arousal scores~\citep{schmidt2018wesad} from the Self-Assessment Manikin (SAM)~\citep{bradley1994sam} -- scored 1 to 9 -- into binary classes using a threshold of 5. Since this threshold is not an established boundary and binarization discards information from the original scale, we retain valence and arousal as regression targets.
\end{enumerate}

This audit informs both which tasks are included in PPG-Bench and how their labels are defined.

\section{Methodology}
	\label{sec:Methodology}
	\paragraph{Four evaluation methods}
	We evaluate each model under four downstream regimes with increasing flexibility. \emph{First}, linear probing uses ridge or logistic regression on frozen embeddings, testing if task-relevant information is directly accessible~\citep{alain2016mlp}. \emph{Second}, a Random Forest allows nonlinear decision boundaries while keeping the encoder fixed. \emph{Third}, a one-layer MLP provides a neural learned nonlinear mapping from the same frozen representation. \emph{Finally}, end-to-end finetuning trains a one-layer MLP jointly with the encoder, allowing the representation itself to adapt to the task. The first three regimes therefore evaluate a fixed pretrained representation with increasing classifier capability, while finetuning additionally measures its ability to adapt.
			
	\paragraph{Controls}
	PPG foundation models should not be evaluated only against one another; a benchmark should also test whether their representations provide value over simpler alternatives. We therefore include three increasingly strong controls.
	
	\begin{enumerate}
		\item \textbf{Random Chance / Constant Baseline} -- A dummy, no-information baseline that predicts the majority training class for classification tasks and the mean training target for regression tasks.
		
		\item \textbf{PRV Control} -- A model that outputs as its embedding standard pulse-rate-variability (PRV) features~\citep{mejia2020prv}, including basic physiological properties such as heart rate and beat-to-beat variability. If an FM does not outperform this control, its learned representational complexity provides no measurable downstream benefit for that task.
		
		\item \textbf{Generic Time-Series FM} -- We compare against MOMENT-1, a general time-series foundation model, to test whether PPG-specific pretraining provides value beyond generic time-series representation learning.
	\end{enumerate}

	Together, these controls ask three questions: whether the PPG signal provides usable predictive information, whether learned FM representations improve over simple physiological features, and whether PPG-specific pretraining improves over a generic time-series foundation model.
		
	\paragraph{Hyperparameter Optimization and Cross-Validation}
	Instead of forcing all models to finetune under the same downstream recipe and hyperparameters, PPG-Bench allows for hyperparameter optimization inside a fairly extensive search space. Models therefore select an optimization best suited to them.
	
	Linear probes and Random Forests use exhaustive hyperparameter grid search. For frozen MLP probes and finetuning, we use Optuna~\citep{akiba2019optuna} with Hyperband~\citep{li2018hyperband}, which allocates larger training budgets only to promising configurations. All search spaces and optimization settings are fixed \textit{a priori} and reported in Appendix~\ref{app:param}.
	
	All evaluations use subject-disjoint 5-fold outer cross-validation. Linear probes and Random Forests use full subject-disjoint inner cross-validation~\citep{cawley2010} to select hyperparameters for each outer fold. Because fully nested evaluation is substantially more expensive for MLP probes and finetuning, these instead use a fixed subject-disjoint validation split within each outer-training fold~\citep{wainer2021nestedcv}.\footnote{For subject-level tasks, we repeat this procedure using three disjoint 20\% validation splits within each outer fold to reduce sensitivity to a single split.}

\section{Results and Analysis}	

\paragraph{Overall Results} We report linear-probe results for the 17 primary tasks in Table~\ref{tab:final-all-methods-17tasks-2x2}, with Random Forest, MLP-probe, and finetuning results in Table~\ref{tab:extended-results}. All evaluations use fully subject-disjoint 5-fold cross-validation over five seeds. Four findings emerge.

\begin{table}[!htbp]
	
	\small
	\setlength{\tabcolsep}{1.4pt}
	\renewcommand{\arraystretch}{1.24}
	\par
	
	\centering
	
	\caption{Benchmark metrics for the 17 primary tasks evaluated under linear probing. Values are mean $\pm$ Student-t 95\% CI. Lower is better for MAE; higher is better for BAC. Bold entries are the best mean and models whose CI overlaps the best.}
	\label{tab:final-all-methods-17tasks-2x2}
	{\normalsize\textbf{Linear Probes}}\par\vspace{2pt}
	\resizebox{\linewidth}{!}{%
		\begin{tabular}{@{}ccc ccccc|ccc@{}}
			\toprule
			Metric & Dataset & Task & AnyPPG & PaPaGei-S & PaPaGei-P & PulsePPG & SigmaPPG & Random & PRV Ctrl & MOMENT \\
			\midrule
			\multirow[c]{10}{*}{MAE} & Dalia & HR & \textbf{7.39}\scoreerr{.18} & 13.75\scoreerr{.32} & 11.68\scoreerr{.31} & 10.00\scoreerr{.27} & 11.28\scoreerr{.30} & 18.23\scoreerr{.42} & 11.19\scoreerr{.34} & 8.24\scoreerr{.19} \\[1.5pt]
			& WildPPG & HR & \textbf{9.76}\scoreerr{.14} & 12.63\scoreerr{.28} & 11.96\scoreerr{.21} & 10.83\scoreerr{.19} & 11.26\scoreerr{.19} & 12.99\scoreerr{.19} & 13.00\scoreerr{.18} & 11.54\scoreerr{.22} \\[1.5pt]
			& \multirow[c]{5}{*}{PPG-BP} & HR & \textbf{3.31}\scoreerr{.06} & 4.89\scoreerr{.15} & 5.22\scoreerr{.16} & 3.67\scoreerr{.15} & 4.15\scoreerr{.10} & 8.63\scoreerr{.02} & 4.48\scoreerr{.12} & \textbf{3.21}\scoreerr{.08} \\
			&  & SBP & \textbf{13.32}\scoreerr{.27} & 14.92\scoreerr{.26} & 14.60\scoreerr{.37} & 13.77\scoreerr{.17} & 14.95\scoreerr{.29} & 16.31\scoreerr{.04} & 16.40\scoreerr{.09} & \textbf{13.89}\scoreerr{.33} \\
			&  & DBP & \textbf{8.21}\scoreerr{.36} & 8.70\scoreerr{.06} & \textbf{8.55}\scoreerr{.13} & \textbf{8.49}\scoreerr{.18} & \textbf{8.63}\scoreerr{.10} & 8.79\scoreerr{.03} & 8.77\scoreerr{.03} & \textbf{8.54}\scoreerr{.27} \\
			&  & Age & 10.76\scoreerr{.24} & 11.37\scoreerr{.21} & 11.54\scoreerr{.36} & \textbf{10.31}\scoreerr{.06} & 11.59\scoreerr{.14} & 12.43\scoreerr{.06} & 12.24\scoreerr{.06} & 10.82\scoreerr{.13} \\
			&  & BMI & \textbf{2.98}\scoreerr{.06} & \textbf{3.00}\scoreerr{.04} & \textbf{3.00}\scoreerr{.05} & \textbf{3.01}\scoreerr{.04} & 3.04\scoreerr{.05} & 3.04\scoreerr{.01} & 3.04\scoreerr{.03} & \textbf{2.97}\scoreerr{.02} \\[1.5pt]
			& BUTPPG & Glucose & 1.31\scoreerr{.08} & 1.24\scoreerr{.08} & 1.23\scoreerr{.08} & 1.26\scoreerr{.10} & 1.22\scoreerr{.09} & 1.23\scoreerr{.07} & 1.23\scoreerr{.12} & 1.30\scoreerr{.10} \\[1.5pt]
			& \multirow[c]{2}{*}{BIDMC} & RR & 2.56\scoreerr{.18} & 2.44\scoreerr{.08} & 2.62\scoreerr{.08} & 2.49\scoreerr{.16} & 2.50\scoreerr{.13} & 2.35\scoreerr{.06} & 2.37\scoreerr{.07} & 2.56\scoreerr{.13} \\
			&  & SpO$_2$ & 2.85\scoreerr{.17} & 2.54\scoreerr{.06} & 2.57\scoreerr{.10} & 2.60\scoreerr{.10} & 2.63\scoreerr{.07} & 2.51\scoreerr{.04} & 2.53\scoreerr{.04} & 2.79\scoreerr{.20} \\
			\addlinespace[4pt]
			\midrule
			\addlinespace[4pt]
			\multirow[c]{7}{*}{BAC} & Dalia & Activity & \textbf{.345}\scoreerr{.007} & .237\scoreerr{.003} & .256\scoreerr{.004} & .294\scoreerr{.005} & .270\scoreerr{.004} & .111\scoreerr{.000} & .239\scoreerr{.002} & \textbf{.346}\scoreerr{.007} \\[1.5pt]
			& WESAD & Emotion & \textbf{.543}\scoreerr{.008} & .396\scoreerr{.008} & .416\scoreerr{.010} & .474\scoreerr{.005} & .456\scoreerr{.005} & .250\scoreerr{.000} & .311\scoreerr{.007} & .504\scoreerr{.015} \\[1.5pt]
			& RealWorldPPG & Identification & .830\scoreerr{.012} & .350\scoreerr{.006} & .385\scoreerr{.010} & \textbf{.909}\scoreerr{.003} & .480\scoreerr{.011} & .029\scoreerr{.000} & .182\scoreerr{.004} & \textbf{.904}\scoreerr{.005} \\[1.5pt]
			& \multirow[c]{2}{*}{PPG-BP} & Diabetes & \textbf{.504}\scoreerr{.012} & .495\scoreerr{.009} & \textbf{.511}\scoreerr{.009} & \textbf{.524}\scoreerr{.035} & \textbf{.511}\scoreerr{.026} & .500\scoreerr{.000} & .500\scoreerr{.000} & \textbf{.536}\scoreerr{.023} \\
			&  & Hypertension & \textbf{.424}\scoreerr{.053} & .306\scoreerr{.023} & \textbf{.356}\scoreerr{.040} & \textbf{.369}\scoreerr{.031} & .332\scoreerr{.031} & .253\scoreerr{.009} & .256\scoreerr{.005} & \textbf{.356}\scoreerr{.039} \\[1.5pt]
			& DREAMT & Sleep Staging & \textbf{.250}\scoreerr{.003} & .244\scoreerr{.001} & \textbf{.246}\scoreerr{.001} & \textbf{.247}\scoreerr{.002} & \textbf{.249}\scoreerr{.001} & .200\scoreerr{.000} & .206\scoreerr{.001} & \textbf{.250}\scoreerr{.001} \\[1.5pt]
			& ECSMP & Depression & .478\scoreerr{.066} & .505\scoreerr{.045} & .499\scoreerr{.049} & .501\scoreerr{.054} & .515\scoreerr{.045} & .520\scoreerr{.056} & .517\scoreerr{.058} & .495\scoreerr{.064} \\
			\bottomrule
		\end{tabular}%
	}
	\par
	
	\par
	
\end{table}

\begin{enumerate}
	\item \textbf{Near-chance performance.}
	On several tasks, no evaluated model meaningfully outperforms the Random Chance / Constant Baseline: 4 of 17 tasks under linear probing, 6 each under Random Forests and frozen MLP probes, and 8 under finetuning. Overall, this occurs in 24/68 method--task combinations (35.3\%).
	
	Several factors may contribute. Some targets, particularly complex diagnostic or demographic labels such as depression, BMI, and diabetes, may contain little information recoverable from a single-channel PPG waveform.\footnote{We nevertheless retain such tasks where they have appeared in prior PPG-FM evaluations.} Other targets are measurable from PPG in principle but may depend on signal characteristics not preserved by current FMs. Conventional SpO$_2$ estimation, for example, uses two simultaneous PPG signals at 2 different wavelengths~\citep{chan2013pulsox}, whereas the evaluated models accept a single waveform and do not explicitly encode wavelength. Respiratory information can appear through low-frequency baseline, amplitude, and frequency modulation; the approximately 0.5\,Hz high-pass filtering used by these models may attenuate some components, given typical resting respiratory frequencies of roughly 0.2--0.33\,Hz~\citep{charlton2018breathing}. Blood pressure and glucose have also been estimated from PPG~\citep{montemoreno2011bpglucose}, but are not reliably recovered by the released FMs here. Finally, the poor MLP and finetuning results on PPG-BP~\citep{liang2018ppgbp} are consistent with overfitting: each participant contributes only three 2.1\,s recordings, leaving little data for high-capacity adaptation.
	
	\item \textbf{Simple physiological features and MOMENT are highly competitive.}
	PPG-specific FMs also frequently fail to outperform the deterministic PRV control, despite PRV using only classical physiological features and no trainable parameters. MOMENT, a generic time-series FM without PPG-specific pretraining, is similarly competitive. A PPG FM outperforms both controls in only 4/17 tasks under linear probing, 4/17 under Random Forests, 2/17 under frozen MLP probing, and 1/17 under finetuning. Thus, in 57/68 method--task combinations (83.8\%), no PPG-specific FM outperforms both controls.
	
	When PRV matches an FM, the FM's learned representation provides no measurable downstream advantage over standard physiological features despite its much greater complexity. Likewise, when MOMENT matches or exceeds the PPG FMs, the released PPG-specific models show no clear advantage from PPG-specific pretraining.\footnote{A causal conclusion about PPG-specific pretraining would require controlled retraining of the same architecture on PPG and generic time-series corpora. Our result concerns the released models as they currently exist.}
	
	\item \textbf{Evaluation methods.}
	Greater downstream flexibility does not uniformly improve performance. Random Forests generally underperform linear probing when FMs are better than chance, while frozen MLP probes often recover or improve upon linear performance (Figure~\ref{fig:transanalysis}). Finetuning further improves several regression tasks but is less consistent for classification.
	
	More importantly, the effect depends on whether the frozen representation is already useful. Among model--task pairs that outperform the random control under linear probing, moving to frozen MLP improves 58.5\% of pairs with a mean gain of 5.9\%, while finetuning improves 78.0\% with a mean gain of 9.3\% (Table~\ref{tab:method-transition-improved-pct}). When performance is no better than the control, added flexibility is usually harmful: linear-to-MLP produces a mean change of $-17.5\%$, and linear-to-finetuning $-29.9\%$. Higher-capacity adaptation therefore helps when useful information is already present, but can catastrophically amplify overfitting when it is not.
	
	\begin{table*}[t]
		\centering
		\caption{Changes between evaluation methods. Improved \% is the fraction of model--task pairs that improve; Mean \% Gain is their average performance change.}
		\label{tab:method-transition-improved-pct}
		\small
		\begin{minipage}[t]{0.49\textwidth}
			\centering
			\textbf{Better than controls}\par\vspace{2pt}
			\resizebox{\linewidth}{!}{%
				\begin{tabular}{@{}lcc@{}}
					\toprule
					Method Transition & Improved \% & Mean \% Gain \\
					\midrule
					Linear Probe $\rightarrow$ Random Forest & 17.1 & -6.7 \\
					Linear Probe $\rightarrow$ Frozen MLP & 58.5 & 5.9 \\
					Linear Probe $\rightarrow$ Finetuning & 78.0 & 9.3 \\
					Frozen MLP $\rightarrow$ Finetuning & 70.7 & 4.5 \\
					\bottomrule
				\end{tabular}%
			}
		\end{minipage}
		\hfill
		\begin{minipage}[t]{0.49\textwidth}
			\centering
			\textbf{Same as controls}\par\vspace{2pt}
			\resizebox{\linewidth}{!}{%
				\begin{tabular}{@{}lcc@{}}
					\toprule
					Method Transition & Improved \% & Mean \% Gain \\
					\midrule
					Linear Probe $\rightarrow$ Random Forest & 41.7 & -3.0 \\
					Linear Probe $\rightarrow$ Frozen MLP & 10.0 & -17.5 \\
					Linear Probe $\rightarrow$ Finetuning & 25.0 & -29.9 \\
					Frozen MLP $\rightarrow$ Finetuning & 55.0 & -12.9 \\
					\bottomrule
				\end{tabular}%
			}
		\end{minipage}
	\end{table*}
	
	\begin{figure}
		\centering
		\includegraphics[width=0.75\linewidth]{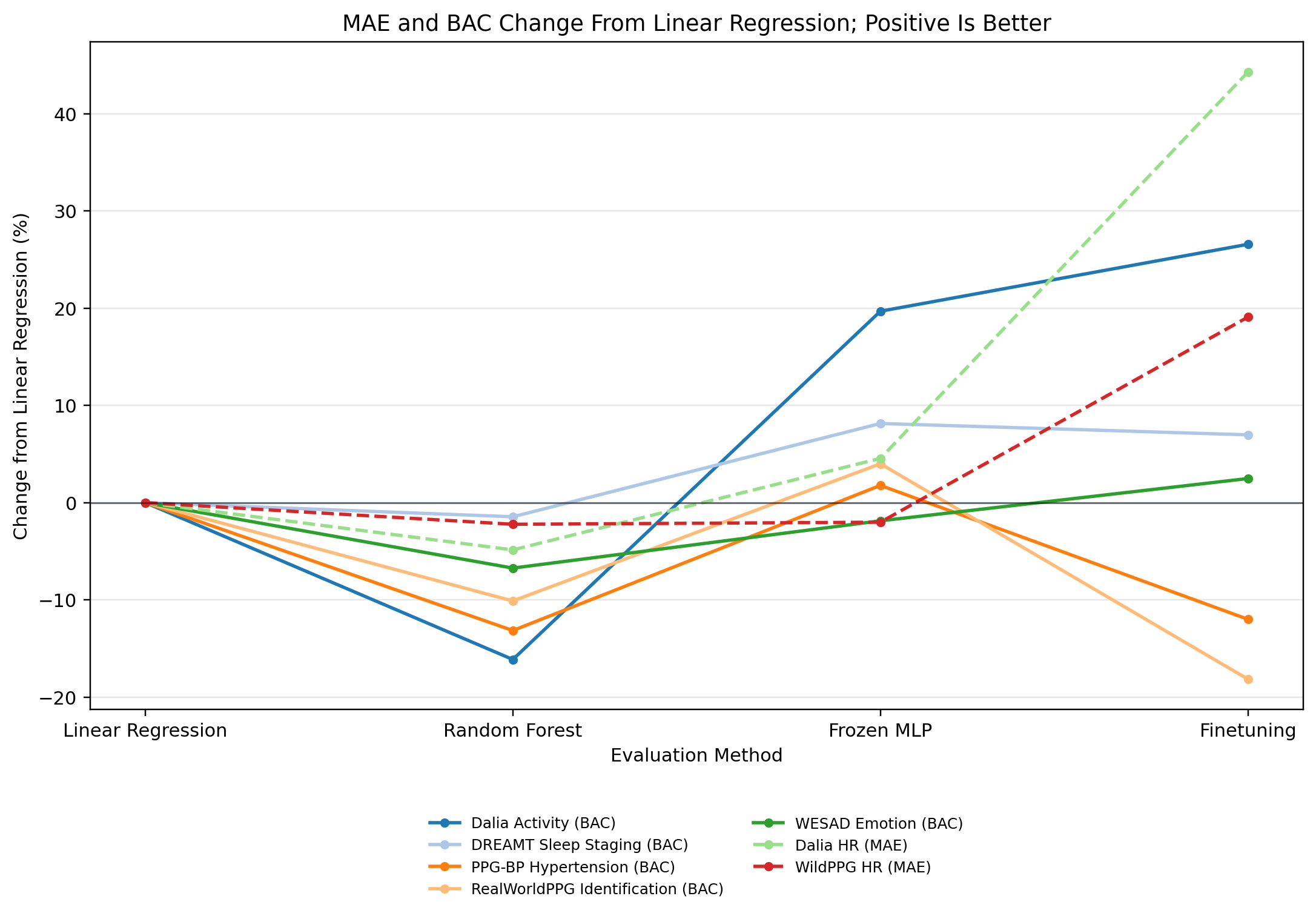}
		\caption{Performance change relative to linear probing for tasks where FMs outperform PRV and the Random Chance / Constant Baseline. Each point is averaged over FMs with above-control performance on that task.}
		\label{fig:transanalysis}
	\end{figure}
	
	\item \textbf{AnyPPG is the strongest PPG-specific model overall.}
	AnyPPG performs best overall among the PPG-specific FMs, with the main exceptions of age estimation and biometric identification. PulsePPG is particularly strong for identification, while MOMENT matches or exceeds PPG-specific models on several tasks. 
\end{enumerate}

Overall, current PPG FMs are not yet consistently strong general-purpose representations: many tasks remain near chance or are matched by PRV or MOMENT, their gains are concentrated in a small subset of tasks, and additional downstream capacity does not reliably rescue weak representations.

\paragraph{Subject-level Evaluation} For tasks in which the target is defined at the participant level, such as diagnostic or other trait-level labels, we evaluate predictions at the subject rather than window level. Although models operate on individual PPG windows, treating multiple windows from the same participant as independent test examples creates a mismatch between the unit of prediction and the unit at which the ground-truth label is defined. We therefore aggregate window-level predictions into a single prediction for each subject before computing evaluation metrics.\footnote{We also treat all PPG-BP targets as subject-level, including HR and blood pressure. Each participant has only three successive 2.1\,s PPG recordings, while each target is provided only once per participant. Given this short recording interval, we treat the three recordings as repeated observations associated with the same ground-truth measurement.}

For example, if a two-minute recording from a subject with a diabetes label produces twelve 10-second windows, the model generates twelve window-level predictions that are aggregated into one diabetes prediction for that subject. We consider three aggregation strategies---mean, median, and top-$k$ mean---with the aggregation strategy selected as a hyperparameter to be selected.

\begin{enumerate}
	\item \textbf{Mean.} For each subject, we take the arithmetic mean of the
	model's prediction vector across all of that subject's windows. This gives
	every window equal weight.
	
	\item \textbf{Median.} For each subject, we take the component-wise median of the model's prediction vector across all windows. Compared with the mean, this is less affected by a small number of noisy or unusual windows.
	
	\item \textbf{Top-$k$ mean.} For each class in a subject's predictions, we set
	$k=\lceil 0.1n\rceil$, where $n$ is the number of windows for the subject.
	We average the $k$ largest scores and the $k$ smallest scores, then retain whichever average is more confident (closer to $1$ or $0$, respectively). This becomes the score for each class. This allows a small, consistently confident subset of windows to determine the subject-level score rather than averaging it away with ambiguous windows.
\end{enumerate}

For subject-level regression, only mean aggregation is used. Top-$k$
mean and median are considered only for classification tasks. Table~\ref{tab:aggregation_selection_frequency_nonregression} details how often models selected the 3 aggregation strategies in classification tasks. We see a fairly even spread across all 4 evaluation methods.

\begin{table}[t]
	\centering
	\caption{Frequency with which each subject-level aggregation method was selected for non-regression tasks in the final benchmark.}
	\label{tab:aggregation_selection_frequency_nonregression}
	\begin{tabular}{lrrr}
		\hline
		Probe method & Mean & Median & Top-$k$ mean \\
		\hline
		Linear probe & 31.8\% & 27.5\% & 40.7\% \\
		Random forest & 34.2\% & 36.2\% & 29.6\% \\
		Frozen MLP & 23.5\% & 34.3\% & 42.2\% \\
		Fine-tuning & 32.5\% & 35.2\% & 32.3\% \\
		\hline
	\end{tabular}
\end{table}

To quantify the effect of performing subject-level aggregation, we rescore the saved pre-aggregation predictions by treating each window as an independent test example, and compare these results with the official subject-level evaluation using the same trained models and predictions. Table~\ref{tab:subject_vs_window_eligible} reports results for tasks where at least one PPG FM outperforms the Random Chance Baseline, averaged across the five PPG FMs. Subject-level aggregation improves performance in every comparison, supporting the principle of evaluation at the same level at which the ground-truth label is defined. Aggregating repeated window predictions also reduces sensitivity to window-level noise, particularly for regression targets.	
\begin{table*}[t]
	\centering
	\small
	\caption{Subject-level versus window-level evaluation for eligible subject-level tasks. Results are averaged across the PPG FMs. Lower is better for MAE; higher is better for BAC. Bold indicates the better score in each row.}
	\label{tab:subject_vs_window_eligible}
	\resizebox{\textwidth}{!}{%
		\begin{tabular}{@{}llcrr@{}}
			\toprule
			Method & Task & Metric & Subject-level & Window-level \\
			\midrule
			\multirow{8}{*}{Linear Probes} & PPG-BP Age & \multirow{5}{*}{MAE} & \textbf{11.117} & 11.714 \\
			& PPG-BP BMI &  & \textbf{3.006} & 3.074 \\
			& PPG-BP DBP &  & \textbf{8.515} & 8.781 \\
			& PPG-BP HR &  & \textbf{4.249} & 5.250 \\
			& PPG-BP SBP &  & \textbf{14.312} & 15.315 \\
			\addlinespace[3pt]
			& PPG-BP Diabetes & \multirow{3}{*}{BAC} & \textbf{0.509} & 0.509 \\
			& PPG-BP Hypertension &  & \textbf{0.357} & 0.335 \\
			& PPG-BP Binary Hypertension &  & \textbf{0.646} & 0.622 \\
			\midrule
			\multirow{6}{*}{Random Forest} & PPG-BP Age & \multirow{4}{*}{MAE} & \textbf{11.602} & 11.859 \\
			& PPG-BP DBP &  & \textbf{8.626} & 8.748 \\
			& PPG-BP HR &  & \textbf{5.147} & 5.476 \\
			& PPG-BP SBP &  & \textbf{15.293} & 15.618 \\
			\addlinespace[3pt]
			& PPG-BP Hypertension & \multirow{2}{*}{BAC} & \textbf{0.308} & 0.303 \\
			& PPG-BP Binary Hypertension &  & \textbf{0.594} & 0.585 \\
			\midrule
			\multirow{5}{*}{Frozen MLP} & PPG-BP Age & \multirow{3}{*}{MAE} & \textbf{11.754} & 12.970 \\
			& PPG-BP HR &  & \textbf{4.773} & 6.203 \\
			& PPG-BP SBP &  & \textbf{15.602} & 17.572 \\
			\addlinespace[3pt]
			& PPG-BP Diabetes & \multirow{2}{*}{BAC} & \textbf{0.524} & 0.518 \\
			& PPG-BP Hypertension &  & \textbf{0.356} & 0.334 \\
			\bottomrule
		\end{tabular}%
	}
\end{table*}

\section{Conclusion}
	We introduce PPG-Bench, a standardized evaluation framework for PPG foundation models spanning diverse datasets, tasks, controls, and downstream adaptation regimes. The benchmark also reveals a clear limitation of current PPG FMs: across many tasks, they do not provide a meaningful advantage over random chance / constant baselines, simple physiological features, or a generic time-series foundation model. Increasing downstream model capacity helps when useful task information is already present, but does not reliably rescue weak representations. These results suggest that current PPG FMs are not yet strong general-purpose representations, and that progress in the field should be judged not only by improvements over other FMs, but by whether those models provide robust value over much simpler alternatives.

\section*{Appendix}
\appendix

	\section{Methodology: Additional Details}
	\subsection{Compute details}
	All evaluations were executed on a cluster of 8 NVIDIA H200 GPUs, except for linear-probe runs, which were performed on CPU. Finetuning was the most computationally expensive evaluation regime (360 GPU-hours), followed by frozen MLP probes (160 GPU-hours) and Random Forests (4 GPU-hours). Linear probes were implemented using \texttt{scikit-learn}~\citep{pedregosa2011sklearn}, Random Forests using \texttt{cuML}~\citep{raschka2020cuml}, and both MLP probing and finetuning using \texttt{PyTorch}~\citep{paszke2019pytorch}. Hyperband HPO was implemented using \texttt{Optuna}~\citep{akiba2019optuna}. PRV features for the control were extracted using NeuroKit2~\citep{makowski2021neurokit}.
	\subsection{Hyperparameter Search Space}
		\label{app:param}
		Table~\ref{tab:hpo-grids} reports the hyperparameter spaces used for each evaluation regime. Linear probes and Random Forests use exhaustive grid search. Frozen MLP probes and finetuning use TPE sampling~\citep{bergstra2011tpe} with Hyperband pruning. Parameters with a single value are fixed rather than searched.
		
		\begin{table*}[t]
			\centering
			\small
			\caption{Hyperparameter search spaces used in PPG-Bench. Parameters with a single value are fixed rather than searched.}
			\label{tab:hpo-grids}
			\begin{tabular}{@{}lll@{}}
				\toprule
				Method & Hyperparameter & Values \\
				\midrule
				
				\multirow{2}{*}{Linear (Ridge)} & $\alpha$ & $\{0.01, 0.1, 1, 10, 100, 1000, 10000\}$ \\
				 & Solver & auto \\
				\addlinespace
				
				\multirow{4}{*}{Linear (Logistic)} & $C$ & $\{0.001, 0.01, 0.1, 1, 10\}$ \\
				 & Penalty & L2 \\
				 & Solver & lbfgs \\
				 & Max iterations & 1000 \\

				\addlinespace
				
				\multirow{6}{*}{Random Forest} & Number of estimators & 100 \\
				 & Max depth & $\{5,10,20\}$ \\
				 & Min. samples split & 2 \\
				 & Min. samples leaf & 1 \\
				 & Max features & sqrt \\

				 & Number of bins & 128 \\
				\addlinespace
				
				\multirow{6}{*}{Frozen MLP} & Hidden dimension & $\{256,512\}$ \\
				 & Dropout & $\{0,0.3\}$ \\
				 & Learning rate & $\{10^{-5},10^{-4},10^{-3},5\times10^{-3},10^{-2}\}$ \\
				 & Weight decay & 0 \\
				 & Batch size & $\{16,32,64,128\}$ \\
				 & Activation & GELU \\
				\addlinespace
				
				\multirow{9}{*}{Finetuning} & Hidden dimension & 256 \\
				 & Head dropout & $\{0,0.3\}$ \\
				 & Head learning rate & $\{10^{-5},10^{-4},10^{-3}\}$ \\
				 & Encoder LR ratio & $\{0.01,0.1\}$ \\
				 & Head weight decay & $\{0,10^{-4}\}$ \\
				 & Encoder weight-decay ratio & 1.0 \\
				 & Batch size & $\{128,512,1024,2048\}$ \\
				 & Activation & GELU \\
				 & BatchNorm mode & \{train, freeze stats\} \\
				\bottomrule
			\end{tabular}
		\end{table*}
		
		For frozen MLP probes and finetuning, we use 48 TPE trials with 8 startup trials. Hyperband uses $r_{\min}=1$ and reduction factor $\eta=3$, with maximum resources of 25 epochs for frozen MLP probes and 15 epochs for finetuning.
		
	\subsection{Edge Pressure.}
		Finite search spaces can constrain performance when selected configurations frequently lie at the boundary of the available grid. We therefore report a post-hoc \textit{edge-pressure} diagnostic for every non-binary numerical HPO axis, together with selected epoch as a diagnostic of the training-budget cap. For each axis, we calculate the fraction of completed evaluations whose selected value lies at the minimum or maximum available setting. This analysis is descriptive and was performed after the benchmark was finalized; it was not used to modify the reported search spaces or results.
		
		\begin{table*}[t]
			\centering
			\small
			\caption{Post-hoc edge pressure in the final hyperparameter searches. Lower- and upper-edge percentages indicate how often the selected value lies at the minimum or maximum available setting. Values above 50\% are bolded.}
			\label{tab:edge-pressure}
			\begin{tabular}{@{}lllrr@{}}
				\toprule
				Method & Hyperparameter & Search range & Lower edge & Upper edge \\
				\midrule
				
				\multirow{2}{*}{Linear} & Logistic $C$ & $[10^{-3}, 10]$ & 14.2\% & 26.9\% \\
				
				 & Ridge $\alpha$ & $[10^{-2}, 10^{4}]$ & 5.8\% & \textbf{50.3\%} \\
				\addlinespace
				
				Random Forest & Max depth & $[5, 20]$ & 43.8\% & 37.9\% \\
				\addlinespace
				
				\multirow{3}{*}{Frozen MLP} & Batch size & $[16, 128]$ & 47.0\% & 13.4\% \\
				
				 & Learning rate & $[10^{-5}, 10^{-2}]$ & 6.0\% & 48.6\% \\
				
				 & Selected epoch & $[1, 25]$ & 10.2\% & 4.6\% \\
				\addlinespace
				
				\multirow{3}{*}{Finetuning} & Batch size & $[128, 2048]$ & \textbf{70.3\%} & 9.1\% \\
				
				 & Head learning rate & $[10^{-5}, 10^{-3}]$ & 7.5\% & \textbf{79.7\%} \\
				
				 & Selected epoch & $[1, 15]$ & 5.1\% & 19.7\% \\
				\bottomrule
			\end{tabular}
		\end{table*}
		
		Table~\ref{tab:edge-pressure} highlights cases where there is significant ($>$50\%) edge pressure. These diagnostics are reported for transparency and to identify search axes that should be expanded in future benchmark reruns.

	\subsection{Model Sizes}
	
	For completeness, we report the approximate parameter counts of all models included in PPG-Bench in Table~\ref{tab:model_sizes}. Model sizes vary substantially, from approximately 4M parameters for AnyPPG to 35.3M for MOMENT-1-small, while the Random Chance and PRV controls are parameter-free. These counts provide additional context when comparing FM performance against the lightweight controls used throughout the benchmark.	
	
	\begin{table}[t]
		\centering
		\caption{Approximate trainable parameter counts for models used in the benchmark.}
		\label{tab:model_sizes}
		\small
		\setlength{\tabcolsep}{6pt}
		\begin{tabular}{lr}
			\toprule
			Model & Trainable Params \\
			\midrule
			PRV Control & 0 \\
			Random & 0 \\
			AnyPPG & 4.0M \\
			PaPaGei-P & 5.0M \\
			PaPaGei-S & 5.8M \\
			PulsePPG & 28.5M \\
			SigmaPPG & 29.7M \\
			MOMENT-1-small & 35.3M \\
			\bottomrule
		\end{tabular}
	\end{table}

	\section{Results, Dataset and Tasks: Additional Details}
		Table~\ref{tab:dataset-overview} details every dataset and task, both primary and secondary, used in this benchmark. Some of the datasets used here are sourced from PhysioNet~\citep{pollard2026physionet}.
		\begin{table*}[t]
			\centering
			\footnotesize
			\caption{Datasets and prediction tasks in the PPG benchmark.}
			\label{tab:dataset-overview}
			\begin{tabularx}{\textwidth}{l X X c c}
				\toprule
				\textbf{Dataset} &
				\textbf{Primary Tasks} &
				\textbf{Secondary Tasks} &
				\textbf{Subjects} &
				\textbf{Recording duration} \\
				\midrule
				Dalia~\citep{reiss2019dalia} &
				HR; activity &
				-- &
				15 &
				36.0 h \\
				
				DREAMT~\citep{wang2024dreamt} &
				Sleep staging &
				-- &
				100 &
				$\sim$700 h \\
				
				WESAD~\citep{schmidt2018wesad} &
				Affective state &
				Valence; Arousal &
				15 &
				$\sim$30 h \\
				
				ECSMP~\citep{gao2021ecsmp} &
				Depression classification &
				Emotion classification &
				67 &
				210.5 h \\
				
				PPG-BP~\citep{liang2018ppgbp} &
				Age; BMI; systolic BP; diastolic BP; HR; diabetes; hypertension &
				Hypertension (binary); cerebral infarction; cerebrovascular disease &
				219 &
				0.383 h \\
				
				BIDMC~\citep{pementel2017bidmc}&
				Respiratory rate; SpO$_2$ &
				-- &
				53 &
				7.07 h \\
				
				BUT-PPG~\citep{nemcova2024butppg} &
				Glucose &
				Systolic BP; diastolic BP; SpO$_2$ &
				50 &
				10.8 h \\
				
				RealWorldPPG~\citep{siam2019realworldppg} &
				Participant identification &
				-- &
				35 &
				3.46 h \\
				
				WildPPG~\citep{meier2024wildppg} &
				HR &
				-- &
				16 &
				216 h \\
				\bottomrule
			\end{tabularx}
		\end{table*}
		
	\subsection{Extended Results}
	We report linear probe results on the primary tasks in Table~\ref{tab:final-all-methods-17tasks-2x2}, and extend them to random forests, MLP probes, and finetuning here.

\begin{table}[!htbp]
\small
\setlength{\tabcolsep}{1.4pt}
\renewcommand{\arraystretch}{1.24}

			\centering
            
            \caption{Benchmark metrics for the 17 primary tasks evaluated with random forests, frozen MLP probes, and finetuning. Values are mean $\pm$ Student-t 95\% CI. Lower is better for MAE; higher is better for BAC. Bold entries are the best mean and models whose CI overlaps the best.}
            \label{tab:extended-results}
			{\normalsize\textbf{Random Forests}}\par\vspace{2pt}
			\resizebox{\linewidth}{!}{%
				\begin{tabular}{@{}ccc ccccc|ccc@{}}
					\toprule
					Metric & Dataset & Task & AnyPPG & PaPaGei-S & PaPaGei-P & PulsePPG & SigmaPPG & Random & PRV Ctrl & MOMENT \\
					\midrule
					\multirow[c]{10}{*}{MAE} & Dalia & HR & \textbf{5.84}\scoreerr{.18} & 14.96\scoreerr{.42} & 13.12\scoreerr{.42} & 10.17\scoreerr{.37} & 13.24\scoreerr{.42} & 18.23\scoreerr{.42} & 11.75\scoreerr{.30} & 8.04\scoreerr{.28} \\[1.5pt]
					& WildPPG & HR & \textbf{9.29}\scoreerr{.10} & 12.55\scoreerr{.26} & 12.45\scoreerr{.23} & 10.99\scoreerr{.22} & 12.16\scoreerr{.21} & 12.99\scoreerr{.19} & 12.78\scoreerr{.19} & 12.06\scoreerr{.23} \\[1.5pt]
					& \multirow[c]{5}{*}{PPG-BP} & HR & \textbf{3.55}\scoreerr{.06} & 5.92\scoreerr{.13} & 6.10\scoreerr{.09} & 5.03\scoreerr{.06} & 5.13\scoreerr{.05} & 8.63\scoreerr{.02} & 4.75\scoreerr{.17} & 4.20\scoreerr{.06} \\
					&  & SBP & \textbf{14.60}\scoreerr{.16} & 15.53\scoreerr{.20} & 15.28\scoreerr{.10} & \textbf{14.87}\scoreerr{.16} & 16.18\scoreerr{.26} & 16.31\scoreerr{.04} & 16.36\scoreerr{.17} & \textbf{14.60}\scoreerr{.21} \\
					&  & DBP & \textbf{8.47}\scoreerr{.19} & \textbf{8.69}\scoreerr{.09} & \textbf{8.63}\scoreerr{.14} & \textbf{8.63}\scoreerr{.08} & \textbf{8.72}\scoreerr{.09} & 8.79\scoreerr{.03} & 8.73\scoreerr{.07} & \textbf{8.53}\scoreerr{.17} \\
					&  & Age & \textbf{11.05}\scoreerr{.16} & 11.92\scoreerr{.13} & 11.83\scoreerr{.07} & \textbf{10.96}\scoreerr{.11} & 12.26\scoreerr{.06} & 12.43\scoreerr{.06} & 12.38\scoreerr{.08} & \textbf{10.94}\scoreerr{.13} \\
					&  & BMI & 3.02\scoreerr{.02} & 3.03\scoreerr{.03} & 2.99\scoreerr{.06} & 3.00\scoreerr{.06} & 3.04\scoreerr{.05} & 3.04\scoreerr{.01} & 3.05\scoreerr{.03} & 2.99\scoreerr{.04} \\[1.5pt]
					& BUTPPG & Glucose & 1.22\scoreerr{.05} & 1.25\scoreerr{.07} & 1.22\scoreerr{.07} & 1.20\scoreerr{.07} & 1.22\scoreerr{.07} & 1.23\scoreerr{.07} & 1.25\scoreerr{.05} & 1.24\scoreerr{.04} \\[1.5pt]
					& \multirow[c]{2}{*}{BIDMC} & RR & 2.53\scoreerr{.11} & 2.40\scoreerr{.07} & 2.44\scoreerr{.06} & 2.47\scoreerr{.10} & 2.45\scoreerr{.10} & 2.35\scoreerr{.06} & 2.52\scoreerr{.14} & 2.57\scoreerr{.15} \\
					&  & SpO$_2$ & 2.76\scoreerr{.15} & 2.55\scoreerr{.06} & 2.52\scoreerr{.07} & 2.70\scoreerr{.15} & 2.60\scoreerr{.06} & 2.51\scoreerr{.04} & 2.55\scoreerr{.06} & 2.74\scoreerr{.12} \\
					\addlinespace[4pt]
					\midrule
					\addlinespace[4pt]
					\multirow[c]{7}{*}{BAC} & Dalia & Activity & \textbf{.290}\scoreerr{.005} & .198\scoreerr{.002} & .218\scoreerr{.003} & .254\scoreerr{.005} & .214\scoreerr{.001} & .111\scoreerr{.000} & .231\scoreerr{.005} & \textbf{.292}\scoreerr{.004} \\[1.5pt]
					& WESAD & Emotion & \textbf{.515}\scoreerr{.011} & .387\scoreerr{.006} & .365\scoreerr{.008} & .465\scoreerr{.010} & .406\scoreerr{.005} & .250\scoreerr{.000} & .359\scoreerr{.014} & .464\scoreerr{.010} \\[1.5pt]
					& RealWorldPPG & Identification & .701\scoreerr{.009} & .439\scoreerr{.006} & .455\scoreerr{.007} & \textbf{.815}\scoreerr{.007} & .322\scoreerr{.005} & .029\scoreerr{.000} & .197\scoreerr{.005} & .736\scoreerr{.006} \\[1.5pt]
					& \multirow[c]{2}{*}{PPG-BP} & Diabetes & .500\scoreerr{.000} & .500\scoreerr{.000} & .497\scoreerr{.002} & .499\scoreerr{.002} & .500\scoreerr{.000} & .500\scoreerr{.000} & .500\scoreerr{.000} & .500\scoreerr{.000} \\
					&  & Hypertension & \textbf{.342}\scoreerr{.035} & .277\scoreerr{.014} & \textbf{.303}\scoreerr{.022} & \textbf{.330}\scoreerr{.022} & .287\scoreerr{.016} & .253\scoreerr{.009} & .270\scoreerr{.026} & \textbf{.332}\scoreerr{.031} \\[1.5pt]
					& DREAMT & Sleep Staging & \textbf{.246}\scoreerr{.001} & \textbf{.245}\scoreerr{.002} & .237\scoreerr{.001} & \textbf{.246}\scoreerr{.002} & \textbf{.245}\scoreerr{.002} & .200\scoreerr{.000} & .206\scoreerr{.001} & \textbf{.247}\scoreerr{.003} \\[1.5pt]
					& ECSMP & Depression & .505\scoreerr{.043} & .515\scoreerr{.061} & .517\scoreerr{.058} & .510\scoreerr{.041} & .517\scoreerr{.058} & .520\scoreerr{.056} & .513\scoreerr{.048} & .507\scoreerr{.044} \\
					\bottomrule
				\end{tabular}%
			}
		\par

\end{table}

\begin{table}[!htbp]
\ContinuedFloat
\caption[]{Continued.}
\small
\setlength{\tabcolsep}{1.4pt}
\renewcommand{\arraystretch}{1.24}

			\centering
			{\normalsize\textbf{Frozen MLP Probes}}\par\vspace{2pt}
			\resizebox{\linewidth}{!}{%
				\begin{tabular}{@{}ccc ccccc|ccc@{}}
					\toprule
					Metric & Dataset & Task & AnyPPG & PaPaGei-S & PaPaGei-P & PulsePPG & SigmaPPG & Random & PRV Ctrl & MOMENT \\
					\midrule
					\multirow[c]{10}{*}{MAE} & Dalia & HR & \textbf{6.12}\scoreerr{.30} & 13.74\scoreerr{.68} & 12.30\scoreerr{1.22} & 9.46\scoreerr{1.11} & 10.39\scoreerr{.79} & 18.23\scoreerr{.42} & 10.35\scoreerr{.80} & 7.49\scoreerr{.05} \\[1.5pt]
					& WildPPG & HR & \textbf{9.89}\scoreerr{.61} & 12.80\scoreerr{1.01} & 12.56\scoreerr{1.44} & \textbf{10.91}\scoreerr{.65} & 11.43\scoreerr{.86} & 12.99\scoreerr{.19} & 13.22\scoreerr{1.09} & 11.75\scoreerr{1.06} \\[1.5pt]
					& \multirow[c]{5}{*}{PPG-BP} & HR & \textbf{3.96}\scoreerr{.11} & 5.20\scoreerr{.39} & 5.68\scoreerr{.16} & \textbf{4.30}\scoreerr{.36} & 4.72\scoreerr{.20} & 8.63\scoreerr{.02} & \textbf{3.94}\scoreerr{.17} & \textbf{4.15}\scoreerr{.63} \\
					&  & SBP & \textbf{14.76}\scoreerr{1.16} & \textbf{15.68}\scoreerr{.94} & \textbf{16.23}\scoreerr{.62} & \textbf{15.01}\scoreerr{.54} & 16.34\scoreerr{.22} & 16.31\scoreerr{.04} & 17.24\scoreerr{1.23} & \textbf{15.51}\scoreerr{.87} \\
					&  & DBP & 9.08\scoreerr{.30} & 9.31\scoreerr{.44} & 9.22\scoreerr{.18} & 9.12\scoreerr{.67} & 9.21\scoreerr{.38} & 8.79\scoreerr{.03} & 9.13\scoreerr{.14} & 9.41\scoreerr{.55} \\
					&  & Age & \textbf{11.22}\scoreerr{.18} & \textbf{11.87}\scoreerr{.23} & 12.07\scoreerr{.16} & \textbf{11.03}\scoreerr{.72} & 12.57\scoreerr{.28} & 12.43\scoreerr{.06} & 12.59\scoreerr{.67} & \textbf{11.35}\scoreerr{.64} \\
					&  & BMI & 3.35\scoreerr{.17} & 3.14\scoreerr{.06} & 3.24\scoreerr{.27} & 3.25\scoreerr{.17} & 3.17\scoreerr{.23} & 3.04\scoreerr{.01} & 3.16\scoreerr{.10} & 3.43\scoreerr{.09} \\[1.5pt]
					& BUTPPG & Glucose & 1.47\scoreerr{.22} & 1.34\scoreerr{.17} & 1.62\scoreerr{.50} & 1.59\scoreerr{.45} & 1.57\scoreerr{.36} & 1.23\scoreerr{.07} & 1.21\scoreerr{.27} & 1.47\scoreerr{.85} \\[1.5pt]
					& \multirow[c]{2}{*}{BIDMC} & RR & 3.09\scoreerr{.44} & 3.01\scoreerr{.30} & 3.40\scoreerr{.75} & 3.17\scoreerr{.25} & 3.20\scoreerr{.14} & 2.35\scoreerr{.06} & 2.63\scoreerr{.44} & 3.48\scoreerr{.20} \\
					&  & SpO$_2$ & 5.87\scoreerr{.85} & 4.14\scoreerr{.67} & 3.87\scoreerr{.97} & 4.57\scoreerr{.67} & 4.30\scoreerr{.51} & 2.51\scoreerr{.04} & 3.16\scoreerr{.26} & 4.94\scoreerr{1.87} \\
					\addlinespace[4pt]
					\midrule
					\addlinespace[4pt]
					\multirow[c]{7}{*}{BAC} & Dalia & Activity & \textbf{.405}\scoreerr{.004} & .289\scoreerr{.023} & .308\scoreerr{.026} & .355\scoreerr{.021} & .338\scoreerr{.006} & .111\scoreerr{.000} & .305\scoreerr{.020} & \textbf{.398}\scoreerr{.030} \\[1.5pt]
					& WESAD & Emotion & \textbf{.516}\scoreerr{.064} & .422\scoreerr{.011} & .418\scoreerr{.025} & \textbf{.467}\scoreerr{.049} & \textbf{.458}\scoreerr{.011} & .250\scoreerr{.000} & .354\scoreerr{.019} & \textbf{.456}\scoreerr{.042} \\[1.5pt]
					& RealWorldPPG & Identification & .829\scoreerr{.036} & .468\scoreerr{.005} & .452\scoreerr{.019} & \textbf{.904}\scoreerr{.011} & .478\scoreerr{.031} & .029\scoreerr{.000} & .185\scoreerr{.016} & \textbf{.881}\scoreerr{.014} \\[1.5pt]
					& \multirow[c]{2}{*}{PPG-BP} & Diabetes & .527\scoreerr{.010} & \textbf{.531}\scoreerr{.022} & \textbf{.501}\scoreerr{.089} & \textbf{.526}\scoreerr{.039} & \textbf{.536}\scoreerr{.053} & .500\scoreerr{.000} & \textbf{.515}\scoreerr{.047} & \textbf{.556}\scoreerr{.012} \\
					&  & Hypertension & \textbf{.411}\scoreerr{.095} & \textbf{.338}\scoreerr{.093} & \textbf{.327}\scoreerr{.091} & \textbf{.382}\scoreerr{.068} & \textbf{.323}\scoreerr{.152} & .253\scoreerr{.009} & .293\scoreerr{.067} & \textbf{.425}\scoreerr{.038} \\[1.5pt]
					& DREAMT & Sleep Staging & \textbf{.270}\scoreerr{.022} & \textbf{.267}\scoreerr{.003} & \textbf{.259}\scoreerr{.009} & \textbf{.271}\scoreerr{.033} & \textbf{.264}\scoreerr{.019} & .200\scoreerr{.000} & .243\scoreerr{.009} & \textbf{.277}\scoreerr{.023} \\[1.5pt]
					& ECSMP & Depression & .479\scoreerr{.086} & .417\scoreerr{.072} & .427\scoreerr{.088} & .455\scoreerr{.153} & .432\scoreerr{.083} & .520\scoreerr{.056} & .504\scoreerr{.076} & .482\scoreerr{.076} \\
					\bottomrule
				\end{tabular}%
			}
		\par

\end{table}

\begin{table}[!htbp]
\ContinuedFloat
\caption[]{Continued.}
\small
\setlength{\tabcolsep}{1.4pt}
\renewcommand{\arraystretch}{1.24}

			\centering
			{\normalsize\textbf{Finetuning}}\par\vspace{2pt}
			\resizebox{\linewidth}{!}{%
				\begin{tabular}{@{}ccc ccccc|ccc@{}}
					\toprule
					Metric & Dataset & Task & AnyPPG & PaPaGei-S & PaPaGei-P & PulsePPG & SigmaPPG & Random & PRV Ctrl & MOMENT \\
					\midrule
					\multirow[c]{10}{*}{MAE} & Dalia & HR & \textbf{5.07}\scoreerr{.29} & 6.21\scoreerr{.40} & 6.51\scoreerr{.18} & \textbf{5.10}\scoreerr{.22} & 6.31\scoreerr{.35} & 18.23\scoreerr{.42} & 10.35\scoreerr{.80} & \textbf{5.52}\scoreerr{.37} \\[1.5pt]
					& WildPPG & HR & \textbf{8.40}\scoreerr{.79} & \textbf{9.27}\scoreerr{.81} & 10.27\scoreerr{.86} & \textbf{8.51}\scoreerr{.46} & \textbf{9.16}\scoreerr{.55} & 12.99\scoreerr{.19} & 13.22\scoreerr{1.09} & \textbf{9.39}\scoreerr{1.78} \\[1.5pt]
					& \multirow[c]{5}{*}{PPG-BP} & HR & \textbf{8.71}\scoreerr{5.41} & 9.91\scoreerr{4.74} & 9.82\scoreerr{4.80} & 14.41\scoreerr{9.71} & 8.75\scoreerr{.12} & 8.63\scoreerr{.02} & \textbf{3.94}\scoreerr{.17} & 15.88\scoreerr{5.07} \\
					&  & SBP & 20.89\scoreerr{10.13} & 16.60\scoreerr{.30} & 22.83\scoreerr{17.33} & 18.87\scoreerr{10.10} & 16.63\scoreerr{.77} & 16.31\scoreerr{.04} & 17.24\scoreerr{1.23} & 28.36\scoreerr{26.14} \\
					&  & DBP & 11.43\scoreerr{5.13} & 8.77\scoreerr{.20} & 12.62\scoreerr{7.20} & 8.76\scoreerr{.16} & 11.16\scoreerr{10.19} & 8.79\scoreerr{.03} & 9.13\scoreerr{.14} & 8.77\scoreerr{.25} \\
					&  & Age & 14.01\scoreerr{3.24} & 12.68\scoreerr{.69} & 14.17\scoreerr{2.45} & 13.69\scoreerr{2.83} & 14.00\scoreerr{2.90} & 12.43\scoreerr{.06} & 12.59\scoreerr{.67} & 14.70\scoreerr{5.13} \\
					&  & BMI & 3.70\scoreerr{.67} & 3.27\scoreerr{.20} & 3.22\scoreerr{.22} & 3.39\scoreerr{.67} & 3.39\scoreerr{.61} & 3.04\scoreerr{.01} & 3.16\scoreerr{.10} & 3.30\scoreerr{.68} \\[1.5pt]
					& BUTPPG & Glucose & 1.40\scoreerr{.56} & 1.28\scoreerr{.28} & 1.36\scoreerr{.29} & 1.27\scoreerr{.44} & 1.25\scoreerr{.40} & 1.23\scoreerr{.07} & 1.21\scoreerr{.27} & 1.25\scoreerr{.47} \\[1.5pt]
					& \multirow[c]{2}{*}{BIDMC} & RR & 2.57\scoreerr{.03} & 2.38\scoreerr{.25} & 2.55\scoreerr{.44} & 2.43\scoreerr{.35} & 2.38\scoreerr{.25} & 2.35\scoreerr{.06} & 2.63\scoreerr{.44} & 2.40\scoreerr{.19} \\
					&  & SpO$_2$ & 3.20\scoreerr{.18} & 2.69\scoreerr{.34} & 2.60\scoreerr{.40} & 2.77\scoreerr{.40} & 2.61\scoreerr{.54} & 2.51\scoreerr{.04} & 3.16\scoreerr{.26} & 2.71\scoreerr{.91} \\
					\addlinespace[4pt]
					\midrule
					\addlinespace[4pt]
					\multirow[c]{7}{*}{BAC} & Dalia & Activity & \textbf{.388}\scoreerr{.015} & .329\scoreerr{.011} & .315\scoreerr{.017} & \textbf{.409}\scoreerr{.010} & .373\scoreerr{.022} & .111\scoreerr{.000} & .305\scoreerr{.020} & \textbf{.399}\scoreerr{.022} \\[1.5pt]
					& WESAD & Emotion & \textbf{.525}\scoreerr{.025} & .437\scoreerr{.011} & .437\scoreerr{.025} & \textbf{.478}\scoreerr{.031} & \textbf{.501}\scoreerr{.053} & .250\scoreerr{.000} & .354\scoreerr{.019} & .480\scoreerr{.010} \\[1.5pt]
					& RealWorldPPG & Identification & .475\scoreerr{.108} & .073\scoreerr{.017} & .075\scoreerr{.030} & \textbf{.927}\scoreerr{.015} & .689\scoreerr{.038} & .029\scoreerr{.000} & .185\scoreerr{.016} & \textbf{.919}\scoreerr{.018} \\[1.5pt]
					& \multirow[c]{2}{*}{PPG-BP} & Diabetes & .513\scoreerr{.028} & \textbf{.499}\scoreerr{.088} & \textbf{.505}\scoreerr{.101} & \textbf{.575}\scoreerr{.098} & \textbf{.548}\scoreerr{.051} & .500\scoreerr{.000} & .515\scoreerr{.047} & \textbf{.605}\scoreerr{.032} \\
					&  & Hypertension & \textbf{.352}\scoreerr{.103} & \textbf{.251}\scoreerr{.029} & \textbf{.302}\scoreerr{.045} & \textbf{.323}\scoreerr{.034} & \textbf{.264}\scoreerr{.033} & .253\scoreerr{.009} & \textbf{.293}\scoreerr{.067} & \textbf{.373}\scoreerr{.097} \\[1.5pt]
					& DREAMT & Sleep Staging & \textbf{.255}\scoreerr{.028} & \textbf{.255}\scoreerr{.008} & \textbf{.260}\scoreerr{.053} & \textbf{.273}\scoreerr{.043} & \textbf{.275}\scoreerr{.020} & .200\scoreerr{.000} & .243\scoreerr{.009} & \textbf{.273}\scoreerr{.015} \\[1.5pt]
					& ECSMP & Depression & .488\scoreerr{.133} & .475\scoreerr{.147} & .499\scoreerr{.126} & .469\scoreerr{.062} & .442\scoreerr{.018} & .520\scoreerr{.056} & .504\scoreerr{.076} & .508\scoreerr{.130} \\
					\bottomrule
				\end{tabular}%
			}
		\par

\end{table}

\subsection{Secondary Tasks}
	\label{sec:secondary}
	We include 9 additional tasks in the benchmark suite for completeness and report preliminary results using linear probes and Random Forests in Table~\ref{tab:linear-rf-extra-tasks}. These tasks either overlap substantially with primary targets or are less commonly used in existing PPG-FM evaluations.
	
	For most secondary tasks, PPG-FM performance does not meaningfully exceed the Random Chance Baseline. The clearest exceptions are WESAD valence and arousal estimation, scored on the 1--9 Self-Assessment Manikin scale~\citep{bradley1994sam}, and PPG-BP binary hypertension classification. The results also continue two trends observed in the primary benchmark: Random Forests generally provide no advantage over linear probing, while AnyPPG is typically the strongest PPG-specific model and is frequently matched by MOMENT.
	\begin{table*}[t]
	\begingroup
	\providecommand{\scoreerr}[1]{{\tiny $\pm$#1}}
	\centering
	\scriptsize
	\setlength{\tabcolsep}{2pt}
	\renewcommand{\arraystretch}{1.24}
	\caption{Metrics for tasks evaluated with linear probes and random forest only. Values are mean $\pm$ Student-t 95\% CI. Lower is better for MAE; higher is better for BAC. Bold entries are the best mean and models whose CI overlaps the best.}
	\label{tab:linear-rf-extra-tasks}
	\vspace{8pt}
	\begin{minipage}[t]{\textwidth}
		\centering
		\textbf{Linear Probe}\par\vspace{2pt}
		\resizebox{\linewidth}{!}{%
			\begin{tabular}{@{}ccc ccccc|ccc@{}}
				\toprule
				Metric & Dataset & Task & AnyPPG & PaPaGei-S & PaPaGei-P & PulsePPG & SigmaPPG & Random & PRV Ctrl & MOMENT \\
				\midrule
				\multirow[c]{5}{*}{MAE} & \multirow[c]{3}{*}{BUTPPG} & SBP & 13.54\scoreerr{.59} & 12.82\scoreerr{.42} & 13.06\scoreerr{.47} & 13.23\scoreerr{.47} & 12.95\scoreerr{.46} & 12.72\scoreerr{.48} & 12.87\scoreerr{.59} & 13.15\scoreerr{.52} \\
				&  & DBP & 6.74\scoreerr{.70} & 6.34\scoreerr{.48} & 6.54\scoreerr{.55} & 6.65\scoreerr{.60} & 6.46\scoreerr{.57} & 6.30\scoreerr{.47} & 6.42\scoreerr{.51} & 6.49\scoreerr{.56} \\
				&  & SpO$_2$ & 1.30\scoreerr{.06} & 1.30\scoreerr{.05} & 1.29\scoreerr{.04} & 1.28\scoreerr{.05} & 1.29\scoreerr{.06} & 1.29\scoreerr{.06} & 1.30\scoreerr{.03} & 1.29\scoreerr{.03} \\[1.5pt]
				& \multirow[c]{2}{*}{WESAD} & Valence & \textbf{0.92}\scoreerr{.02} & 1.12\scoreerr{.02} & 1.10\scoreerr{.05} & 1.00\scoreerr{.04} & 1.08\scoreerr{.03} & 1.17\scoreerr{.02} & 1.16\scoreerr{.03} & 1.00\scoreerr{.05} \\
				&  & Arousal & \textbf{1.29}\scoreerr{.03} & 1.61\scoreerr{.02} & 1.67\scoreerr{.02} & 1.46\scoreerr{.03} & 1.53\scoreerr{.03} & 1.78\scoreerr{.02} & 1.78\scoreerr{.01} & \textbf{1.36}\scoreerr{.05} \\[1.5pt]
				\addlinespace[4pt]
				\midrule
				\addlinespace[4pt]
				\multirow[c]{4}{*}{BAC} & \multirow[c]{3}{*}{PPG-BP} & Hypertension (Binary) & \textbf{.691}\scoreerr{.032} & .620\scoreerr{.030} & .633\scoreerr{.011} & \textbf{.674}\scoreerr{.024} & .609\scoreerr{.016} & .500\scoreerr{.000} & .499\scoreerr{.002} & \textbf{.653}\scoreerr{.031} \\
				&  & Cerebral Infarction & .525\scoreerr{.055} & .516\scoreerr{.055} & .517\scoreerr{.057} & .531\scoreerr{.082} & .511\scoreerr{.046} & .520\scoreerr{.056} & .519\scoreerr{.053} & .518\scoreerr{.057} \\
				&  & Cerebrovascular Disease & .509\scoreerr{.030} & .498\scoreerr{.003} & .500\scoreerr{.008} & .498\scoreerr{.015} & .494\scoreerr{.004} & .500\scoreerr{.000} & .500\scoreerr{.000} & .503\scoreerr{.024} \\[1.5pt]
				& ECSMP & Emotion & .162\scoreerr{.005} & .165\scoreerr{.005} & .166\scoreerr{.003} & .168\scoreerr{.003} & .168\scoreerr{.007} & .167\scoreerr{.000} & .170\scoreerr{.005} & .166\scoreerr{.003} \\
				\bottomrule
			\end{tabular}%
		}
	\end{minipage}
	\par\vspace{2.0em}
	\begin{minipage}[t]{\textwidth}
		\centering
		\textbf{Random Forest}\par\vspace{2pt}
		\resizebox{\linewidth}{!}{%
			\begin{tabular}{@{}ccc ccccc|ccc@{}}
				\toprule
				Metric & Dataset & Task & AnyPPG & PaPaGei-S & PaPaGei-P & PulsePPG & SigmaPPG & Random & PRV Ctrl & MOMENT \\
				\midrule
				\multirow[c]{5}{*}{MAE} & \multirow[c]{3}{*}{BUTPPG} & SBP & 13.75\scoreerr{.60} & 12.93\scoreerr{.49} & 13.07\scoreerr{.42} & 13.55\scoreerr{.56} & 12.97\scoreerr{.50} & 12.72\scoreerr{.48} & 13.42\scoreerr{.73} & 13.59\scoreerr{.53} \\
				&  & DBP & 6.78\scoreerr{.71} & 6.45\scoreerr{.51} & 6.50\scoreerr{.56} & 6.76\scoreerr{.68} & 6.48\scoreerr{.54} & 6.30\scoreerr{.47} & 6.54\scoreerr{.67} & 6.56\scoreerr{.63} \\
				&  & SpO$_2$ & 1.29\scoreerr{.05} & 1.32\scoreerr{.06} & 1.29\scoreerr{.05} & 1.29\scoreerr{.05} & 1.29\scoreerr{.06} & 1.29\scoreerr{.06} & 1.32\scoreerr{.04} & 1.29\scoreerr{.04} \\[1.5pt]
				& \multirow[c]{2}{*}{WESAD} & Valence & \textbf{0.90}\scoreerr{.02} & 1.13\scoreerr{.03} & 1.11\scoreerr{.03} & 1.00\scoreerr{.03} & 1.10\scoreerr{.02} & 1.17\scoreerr{.02} & 1.09\scoreerr{.02} & 1.00\scoreerr{.02} \\
				&  & Arousal & \textbf{1.31}\scoreerr{.03} & 1.64\scoreerr{.02} & 1.66\scoreerr{.03} & 1.47\scoreerr{.03} & 1.59\scoreerr{.02} & 1.78\scoreerr{.02} & 1.62\scoreerr{.01} & 1.47\scoreerr{.02} \\[1.5pt]
				\addlinespace[4pt]
				\midrule
				\addlinespace[4pt]
				\multirow[c]{4}{*}{BAC} & \multirow[c]{3}{*}{PPG-BP} & Hypertension (Binary) & \textbf{.627}\scoreerr{.038} & \textbf{.567}\scoreerr{.031} & \textbf{.592}\scoreerr{.024} & \textbf{.622}\scoreerr{.028} & \textbf{.561}\scoreerr{.031} & .500\scoreerr{.000} & .503\scoreerr{.007} & \textbf{.620}\scoreerr{.023} \\
				&  & Cerebral Infarction & .520\scoreerr{.056} & .520\scoreerr{.056} & .520\scoreerr{.056} & .520\scoreerr{.056} & .520\scoreerr{.056} & .520\scoreerr{.056} & .520\scoreerr{.056} & .520\scoreerr{.056} \\
				&  & Cerebrovascular Disease & .500\scoreerr{.000} & .500\scoreerr{.000} & .500\scoreerr{.000} & .500\scoreerr{.000} & .500\scoreerr{.000} & .500\scoreerr{.000} & .500\scoreerr{.000} & .500\scoreerr{.000} \\[1.5pt]
				& ECSMP & Emotion & .163\scoreerr{.002} & .170\scoreerr{.004} & .167\scoreerr{.003} & .163\scoreerr{.005} & .167\scoreerr{.005} & .167\scoreerr{.000} & .166\scoreerr{.002} & .158\scoreerr{.003} \\
				\bottomrule
			\end{tabular}%
		}
	\end{minipage}
	\endgroup
\end{table*}

\end{document}